\pdfoutput=1 
\documentclass[runningheads]{llncs}
\usepackage[T1]{fontenc}
\usepackage{graphicx}
\usepackage{amsmath}
\usepackage{eucal}
\usepackage{amsfonts}
\usepackage{booktabs}
\usepackage{multirow}
\usepackage{textcomp}
\usepackage{amssymb}
\usepackage{graphicx}
\usepackage{bibnames}
\usepackage{breqn}
\usepackage{csquotes}
\usepackage{siunitx}
\usepackage{url}
\usepackage{color}

\begin{document}
\title{Fuzzy Logic Visual Network (FLVN): A neuro-symbolic approach for visual features matching }
\titlerunning{FLVN:  A neuro-symbolic approach for visual features matching}
%

\author{Francesco Manigrasso\inst{1}\orcidID{0000-0002-4151-8880} \and
Lia Morra\inst{1}\orcidID{0000-0003-2122-7178} \and
Fabrizio Lamberti\inst{1}\orcidID{0000-0001-7703-1372}}
\authorrunning{F. Manigrasso et al.}
%
\institute{Politecnico di Torino, Dipartimento di Automatica e Informatica, Torino, Italy\\
\email\{francesco.manigrasso, lia.morra, fabrizio.lamberti\}@polito.it}

\maketitle              
\begin{abstract}
Neuro-symbolic integration aims at harnessing the power of symbolic knowledge representation combined with the learning capabilities of deep neural networks. In particular, Logic Tensor Networks (LTNs) allow to incorporate background knowledge in the form of logical axioms by grounding a first order logic language as differentiable operations between real tensors. Yet, few studies have investigated the potential benefits of this approach to improve zero-shot learning (ZSL) classification. In this study, we present the Fuzzy Logic Visual Network (FLVN) that formulates the task of learning a visual-semantic embedding space within a neuro-symbolic LTN framework. FLVN incorporates prior knowledge in the form of class hierarchies (classes and macro-classes) along with robust high-level inductive biases. The latter allow, for instance, to handle exceptions in class-level attributes, and to enforce similarity between images of the same class, preventing premature overfitting to seen classes and improving overall performance.  FLVN reaches state of the art performance on the Generalized ZSL (GZSL) benchmarks AWA2 and CUB, improving by 1.3\% and 3\%, respectively. Overall, it achieves competitive performance to recent ZSL methods with less computational overhead.
FLVN is available at
\url{https://gitlab.com/grains2/flvn}.

\end{abstract}

\makeatletter
\let\oldabs\abs
\def\abs{\@ifstar{\oldabs}{\oldabs*}}
\let\oldnorm\norm
\def\norm{\@ifstar{\oldnorm}{\oldnorm*}}
\makeatother

\newcommand{\XX}{\mathcal{X}}
\newcommand{\CC}{\mathcal{C}}
\newcommand{\FF}{\mathcal{F}}
\newcommand{\PP}{\mathcal{P}}
\newcommand{\GG}{\mathcal{G}}
\newcommand{\DD}{\mathcal{D}}
\newcommand{\LL}{\mathcal{L}}
\newcommand{\KK}{\mathcal{K}}
\newcommand{\prob}{\mathbb{P}}
\newcommand{\reals}{\mathbb{R}}
\newcommand{\Din}{D_{\mathrm{in}}}
\newcommand{\Dout}{D_{\mathrm{out}}}
\newcommand{\naturals}{\mathbb{N}}
\newcommand{\sigmoid}{\mathrm{sigmoid}}
\newcommand{\fuz}{\mathrm{FuzzyOp}}
\newcommand{\satagg}{\mathrm{SatAgg}}
\newcommand{\argmaxx}{\mathrm{argmax}}
\newcommand{\argminn}{\mathrm{argmin}}

\section{Introduction}
There is an increasing interest in the computer vision community towards the integration of learning and reasoning, with specific emphasis towards the fusion of deep neural networks and symbolic artificial intelligence \cite{Yu2021RecentAI,LTN,van2021modular}. By integrating perception, reasoning and learning, Neuro-symbolic (NeSy) architectures can improve explainability, generalization and robustness of deep learning systems. In addition, NeSy techniques can incorporate prior symbolic knowledge in the training objective of deep neural networks, compensating for the lack of supervision from labelled examples \cite{manigrasso2021faster,donadello2019compensating}.  The latter property is of particular interest in the context of Zero-Shot Learning (ZSL), which aims at recognizing objects from unseen classes by associating both seen and unseen classes to auxiliary information, usually in the form of class attributes \cite{awa2}.

In this paper, we introduce the Fuzzy Logic Visual Network (FLVN), an architecture designed to learn a joint embedding space for visual features and class attributes to tackle ZSL scenarios. To constraint learning based on prior knowledge available from resources such as WordNet, it leverages the Logic Tensor Network (LTN)\cite{LTN} paradigm, a NeSy framework that formulates learning as maximizing the satisfiability of a knowledge base $\mathcal{K}$ based on a first order logic (FOL). Thus, FLVN is composed by a convolutional neural network (CNN), that computes embedding, followed by a LTN.

The proposed FLVN improves and extends previous NeSy architectures for ZSL, and in particular our previous architecture ProtoLTN \cite{Martone2022PROTOtypicalLT}, in several ways. First, there are substantial differences in the LTN formulation, and especially the way the isOfClass predicate is formulated. FLVN, learns to project visual features into the semantic attribute space, facilitating the integration of prior knowledge into the learning process by the LTN. FLVN is trained end-to-end, whereas ProtoLTN used features extracted from a pre-trained frozen network, and only the class-level prototypes were trained. This implementation outperforms our previous model \cite{Martone2022PROTOtypicalLT}, offering an alternative approach for grounding the predicates in the FOL language by changing the how the distance measure between extracted features and semantic vectors is calculated. 

Compared to most existing approaches to ZSL, FLVN incorporates into the training process several logical axioms that combine both class-specific prior knowledge (e.g., ``a zebra is an ungulate'') with high-level inductive biases (e.g., ``two images of the same class should have similar embeddings''), that were not yet included in ProtoLTN \cite{Martone2022PROTOtypicalLT}. Finally, we introduce axioms to represent exceptions within the dataset to make the architecture more robust in the context of ZSL,  where class attributes are specified at the class level rather than the image level. To the best of our knowledge, this aspect is not taken into account in current approaches to ZSL, including neuro-symbolic ones \cite{Martone2022PROTOtypicalLT}. Through extensive experiments on multiple benchmarks, we show that FLVN outperforms existing ZSL architectures.  The implementation, developed in PyTorch using the LTNtorch package \cite{LTNtorch}, is available at
\url{https://gitlab.com/grains2/flvn}.

The remainder of the paper is structured as follows. Section \ref{sec:related} introduces background on the LTN framework and related work. Section \ref{sec:flvn} describes the proposed architecture. In Sections \ref{sec:Experiments} and \ref{sec:Results}, we analyze the model behavior in a ZSL and a generalized ZSL (GZSL) setting on common benchmarks. Finally, in Section \ref{sec:Conclusion}, we discuss the results and future work.

\section{Related work}
\label{sec:related}

\subsection{Neural-symbolic AI in semantic image interpretation}
In recent years, there has been significant research focus on NeSy architectures for addressing semantic image interpretation tasks \cite{Yu2021RecentAI,LTN,manigrasso2021faster,donadello2019compensating,Martone2022PROTOtypicalLT,li2021calibrating}. 
The present studies falls within the class of NeSy
techniques that seeks to incorporate symbolic information as a prior \cite{van2021modular}. Specifically, we rely on LTNs, which are modular architectures capable of incorporating FOL constraints \cite{LTN,Martone2022PROTOtypicalLT} and can be jointly trained with  neural module in an end-to-end manner \cite{manigrasso2021faster}.

In the LTN framework, the concept of  \textit{grounding} plays a crucial role in interpreting FOL within a specific subset of the domain, denoted as $\mathbb{R}^{n}$. In this approach, logical predicates and axioms are represented as vectors, which are then grounded (interpreted) as real numbers in the range [0, 1] using a technique called  \textit{Real Logic}. By employing this grounding mechanism, the LTN framework maps each term, denoted as $x$, to a vector representation in $\mathcal{G}(x) = \mathbb{R}^{n}$, while each predicate symbol, represented by $p \in \mathcal{P}$, is mapped to $\mathcal{G}\left(D(p)\right) \rightarrow [0,1]$. The concept can be illustrated by considering the frequently encountered \texttt{isOfClass} predicate in neuro-symbolic architectures, which quantifies the probability of a given term belonging to a specific class $c$ \cite{manigrasso2021faster,Martone2022PROTOtypicalLT}. The training objective is formulated by constructing a knowledge base $\mathcal{K}$ of FOL axioms, and finding the \textit{best satisfiability} (sat). 

\subsection{Zero-shot learning}
ZSL tasks entail recognizing objects from previously unseen classes by exploiting some form of auxiliary knowledge, usually attribute-based, learned from seen classes. GZSL extends ZSL by assuming both seen and unseen classes are present at test time. Different strategies have been proposed to tackle ZSL, including embedding-based, attention mechanism-based, and generative strategies.

\textbf{Embedding-based} techniques compare semantic attributes and visual information by mapping them onto a suitable embedding space. Some methods embed images into the attribute space using an embedding function and consider semantic attributes as the common space~\cite{APN,CC-ZSL}. Other techniques have proposed to adopt the image embedding space as common space~\cite{Martone2022PROTOtypicalLT,dem,vse} to mitigate the hubness problem~\cite{Zhang2016LearningAD}. 
A third class of approaches~\cite{SP-AEN,EDEZSL,TCN} utilizes a shared space distinct from both image and attribute domains. To prevent overfitting to seen classes, these approaches require the use of pseudo-labeling techniques or a transductive setting~\cite{EDEZSL}, assuming that unlabeled images from unseen classes are provided during training. \textbf{Attention mechanisms} have been proposed to find the image regions that contribute the most to the categorization of a certain class and improve the embedding space~\cite{APN,CC-ZSL,Li2021AttributeModulatedGM,AREN}.

A critical aspect of embedding-based method is preventing overfitting to seen classes using, e.g., regularization \cite{APN,EDEZSL} or contrastive techniques \cite{CC-ZSL}. 
FLVN accomplishes this objective by incorporating a symbolic prior to aggregate visually and semantically similar features, while also explicitly establishing relationships between classes within the same macro class (a characteristic often implicitly addressed in alternative methodologies).

Previous approaches have used semantic class descriptions for the classification, but attributes linked to a specific class may not be consistently expressed or detectable in individual images. FLVN addresses this limitation by incorporating axioms that encode the existence of ``exceptions to the rule'' within the dataset. This aspect, which has received limited consideration in previous studies, has been experimentally shown to improve classification accuracy.

\textbf{Generative techniques}, on othe other hand, exploit auxiliary models, such as Generative Adversarial Networks (GANs), to generate artificial examples representing unseen classes by learning a conditional probability for each class~\cite{E-PGN,TGMZ,LisGAN,cycle-CLSWGAN}. Recently, feature generation models were integrated with embedding-based models in a contrastive setting~\cite{CEGZSL}. Generative methods require prior knowledge regarding unseen classes when generating training data. In contrast, the training process of FLVN relies on a subset of distinct seen classes and solely assumes knowledge of the class hierarchy at training time, so that the training can be easily extended to new unseen classes. Nonetheless, our approach is complementary to generative techniques, and could be in principle combined.

\section{Fuzzy Logic Visual Network}
\label{sec:flvn}
\begin{figure*}[tb!]
\centering
\includegraphics[scale=0.6]{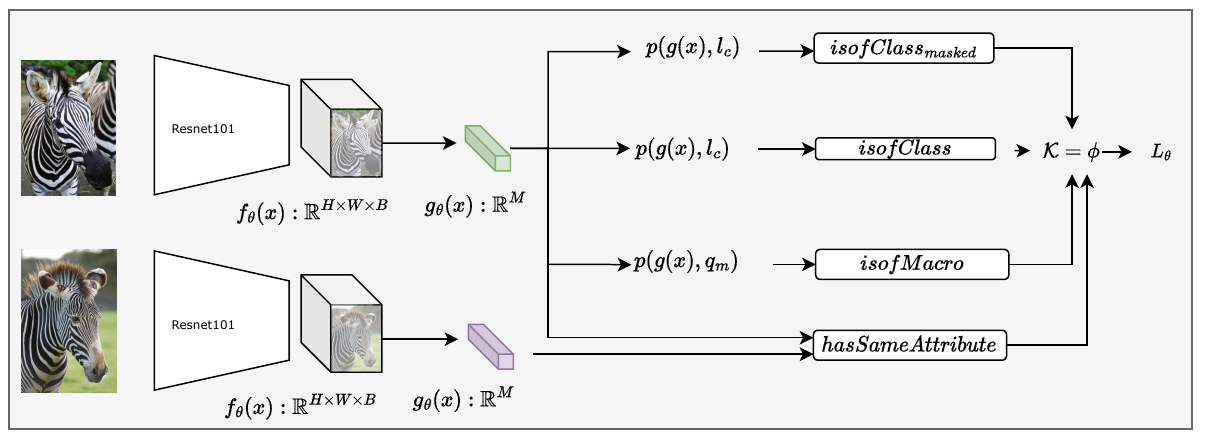}
\caption{The FLVN architecture, designed for Zero-Shot Learning (ZSL) classification, combines a convolutional feature extractor and an attribute encoder to effectively map visual features to the attribute space. By leveraging predicates such as \texttt{isOfClass}, $\texttt{isOfClass}_{\text{masked}}$, and \texttt{isOfMacro}, it reduces the disparity between image features and class attributes, while the \texttt{hasSameAttribute} predicate captures the similarity between two images. All these predicates play a crucial role in the formulation of the architecture, as they are incorporated within the formulas of a knowledge base denoted as $\KK$. During the training process, the loss function is designed to optimize the satisfiability (truth value) of all the formulas within $\KK$.}
\label{Figure:architecture}
\end{figure*}

We introduce a comprehensive and trainable framework for the task of ZSL, depicted in Figure~\ref{Figure:architecture}. 
It comprises two main modules: a feature extractor and a LTN that formulates the training objective.
\subsubsection{Feature extractor.} 
 The feature (embedding) extractor is a CNN that maps the input $x$ to a feature space $f_{\theta}(x) \in \mathbb{R} ^{H \times W \times B}$, where $H$, $W$ and $B$ represent the height, width, and number of channels of the features, respectively. Through mean pooling over $H$ and $W$, we obtain global discriminative characteristics $g_{\theta}(x) \in \mathbb{R}^{B \times 1}$, and utilize a linear projection to transform them into a semantic space represented by $V \in \mathbb{R}^{B \times M}$, where $B$ is the dimension of the vector space of features, and $M$ denotes the length of the attribute vector \cite{APN}.

\subsubsection{Logic Tensor Network.}
The LTN  formulates the learning objective as the maximum satisfiability of a $\KK$ .  For each training batch, the $\KK$ is updated introducing axioms that represent labelled examples ($\phi_1$), as well as prior knowledge ($\phi_2, \phi_3, \phi_4, \phi_5,\phi_6$). The maximum satisfiability loss is then defined based on the aggregation of all axioms as follows:
\begin{equation}
        \mathcal{L}^{\text{ep}} = 1 -\left(\bigwedge_{\phi \in \mathcal{K}} \phi \right)  = 1 -\GG(\phi) 
        \label{eq:loss}
\end{equation}

This section first define the variables, predicates and domain that form the FOL language, followed by the definition of the knowledge base $\KK$. 

\textbf{Groundings}. Variables and their domains are grounded as follows:

\begin{equation}
\GG(l)= \mathbb{N}^{C}, 
\GG(q)= \mathbb{N}^{Q}
\end{equation}
\begin{equation}
\GG(a)=  \GG(a^{\text{mask}})= \mathbb{R}^{M \times C} 
\end{equation}
\begin{equation}
\GG(a^{\text{macro}})= \mathbb{R}^{M \times Q} 
\end{equation}
\begin{equation}
\GG(x)=  g_\theta(f_\theta(\GG(\texttt{images}))) = \mathbb{R}^{M}  \\
\end{equation}
where the variable $l$ represents the class labels belonging to set of classes $C$, $q$ represent the macroclass label belonging to set of macroclasses $Q$, and each class/macroclass is described by a set of non-binary semantic attributes denoted by $a$ and $a^{\text{macro}}$, respectively. Functions $f_\theta$ and $g_\theta$ are employed to embed images into the attribute space, resulting in the final representation $\GG(x)$. The FOL language contains four main predicates: $\texttt{isOfClass}(x,l)$ and $\texttt{isOfClass}_{\text{masked}}(x,l)$ denote the fact that an image $x$ belongs to class $l$,
$\texttt{isOfMacro}(x,q)$ that an image $x$ belongs to the macroclass $q$, and $\texttt{hasSameAttribute}(x_1, x_2)$ that two images have the same attributes. 

The $\GG(\texttt{isOfClass})$ predicate is grounded by the similarity  between the input image and the corresponding class attribute vectors. First, we compute the similarity between the image $x$ and a class $l_c$ by calculating the scaled product of the global features mapped in the attribute space with the semantic vectors:
\begin{equation}
p\left(x,l\right)=\frac{\exp \left(x^T V a_{l} \right)}{\sum_{s=1}^S \exp \left(x^T V a_s \right)}
\label{eq:prob}
\end{equation}
where $a_{y}$ represents the semantic attribute vector associated with class $l$. To obtain a prediction score for an example $x$, we calculate the dot product between the output of $p$ (Eq. \ref{eq:prob}) and the one-hot encoding $l_{c}^T$ for class $c \in C$, as follows:

\begin{equation}
\GG(\texttt{isOfClass}):x,l_c \rightarrow {l_c}^T p(\GG(x),l_c)
\end{equation}

 Similarly, we define $\GG(\texttt{isOfMacro})$ and $\GG(\texttt{isOfClass}_{\text{masked}})$. Since attributes for macro-classes are in principle unknown, we define a trainable attribute vector $a_m^{\text{macro}}$ for macro-class $q_m$ to compute $\GG(\texttt{isOfMacro})$. On the other hand, $\GG(\texttt{isOfClass}_{\text{masked}})$ uses the masked attribute vector $a_c^{\text{masked}}$ for class $l_c$, in which missing attributes $k$ are set to 0, while preserving the rest of the attributes in $a_c$. Finally, the grounding for $\texttt{hasSameAttribute}$  is defined as:
\begin{equation}
\GG(\texttt{hasSameAttribute}):x_1,x_2 \rightarrow   \texttt{sigmoid}(\alpha  \texttt{d}(\GG(x_1),\GG(x_2))) 
\label{eq:gt_sameattributes}
\end{equation}
where $\texttt{d}$ is the cosine similarity, $\alpha$ a scale factor and $\GG(x_1)$, $\GG(x_2)$ correspond to the embeddings of the two images.

\textbf{Learning from labeled examples}. 
We incorporate labelled examples by introducing an axiom $\phi_1$ stating that all facts about labeled example should be true, that is, all labeled samples should be classified correctly:
\begin{equation}
\phi_{\text{1}} = \forall \text{Diag}(x,l_c) (\texttt{isOfClass}(x,l_c))   \label{eq:isOfClass} \footnote{Diagonal Quantification quantifies over pairs of instances, e.g., images and their labels. A more formal definition can be found in \cite{LTN}.}
\end{equation}

To account for the class hierarchy, we also introduce an axiomatic statement $\phi_2$ to indicate that ``if an image contains a zebra'', then ``the image belongs to the family of ungulates'':

\begin{equation}
\phi_{\text{2}} = \forall \text{Diag}(x,l_c,q_m) (\texttt{isOfClass}(x,l_c) \implies \texttt{isOfMacro}(x,q_m)) 
\label{eq:isOfMacro}
\end{equation}

\textbf{Learning better feature representations}. The following axiom encodes the assumption that features extracted from two images of the same class should possess the same attributes:

\begin{equation}
\phi_{\text{3}} = \forall \text{Diag}(x_1,l_{c_1}) \bigg( \forall\text{Diag}(x_2, l_{c_2} ):{c_1}={c_2} \texttt{ }
 \texttt{hasSameAttribute}(x_1,x_2) \bigg)
 \label{eq:sameattribute}
\end{equation}

Likewise, images from different classes should possess different attributes:
\begin{equation}
\phi_{\text{4}} = \forall \text{Diag}(x_1,l_{c_1}) \bigg(\forall\text{Diag}(x_2, l_{c_2} ):{c_1}!={c_2} \texttt{ }
 \neg\texttt{hasSameAttribute}(x_1,x_2)\bigg)
 \label{eq:notsameattribute}
\end{equation}

To further emphasize the similarity between visual attributes and semantic vectors, the following axiom enforces the similarity between image embeddings and attribute vectors of the same class:
\begin{equation}
\phi_{\text{5}} = \forall \text{Diag}(x,l_{c}) \bigg( \forall\text{Diag}(a, l_{a} ): c=a
 \texttt{ hasSameAttribute}(x,a) \bigg)
 \label{eq:sameattribute_class}
\end{equation} 

\textbf{Learning with refutation}. In classical ZSL benchmarks such as AWA2, attributes are associated with class labels using a crisp or fuzzy matrix. However, this association does not entail that all examples of a class will exhibit exactly the same attributes: attributes may be expressed by a subset of training samples, or may be occluded. The existential statement $\phi_6$ represents the fact that some class attributes may not be present for all samples (e.g, ``there exists a zebra that is not agile''). Given that image-level attributes are not available, we simply remove randomly selected attributes by defining the $\texttt{isOfClass}_{\text{masked}}$ predicate:
\begin{equation}
\phi_{\text{6}} = \forall  l^{\text{seen}}
( \exists x , \texttt{isOfClass}_{\text{masked}}(x,l^{\text{seen}})  )
\label{eq:isOfClassmasked}
\end{equation}
where $l^{\text{seen}}$ denotes the list of seen classes.

\textbf{Grounding logical connectives and aggregators}.
The knowledge base $\KK$ is an aggregation of formulas  updated at each training step. To solve the maximum satisfiability problem using gradient descent, logical connectives and aggregators must be grounded into Real Logic. Given  two truth values $a$ and $b$ in $[0,1]$, we adopted the symmetric configuration from \cite{LTN}, using the standard negation $\neg: N_S(a) =1-a $ and the Reichenbach implication $\rightarrow: I_R(a, b)  =1-a+a b$. The existential quantifier $\exists$ was approximated by the generalized mean $A_{pM}$, and the universal quantifier $\forall$ by the generalized mean w.r.t. the error $A_{pME}$, respectively \cite{LTN,van2008visualizing}.  Given $n$ truth values $a_1, \ldots, a_n$ all in $[0,1]$:
\begin{equation}
 \exists: A_{p M}\left(a_1, \ldots, a_n\right)=\left(\frac{1}{n} \sum_{i=1}^n a_i^{p_{\exists}}\right)^{\frac{1}{p_{\exists}}} \quad p_{\exists} \geqslant 1 \\
 \label{eq:aggregmean}
\end{equation}
\begin{equation}
\forall: A_{p M E}\left(a_1, \ldots, a_n\right)=1-\left(\frac{1}{n} \sum_{i=1}^n\left(1-a_i\right)^{p_{\forall}}\right)^{\frac{1}{p_{\forall}}} \quad p_{\forall} \geqslant 1\\
\label{eq:aggregmeanerror}
\end{equation}

$A_{p M E}$ is a measure of how much, on average, truth values $a_i$ deviate from the true value of 1. The $A_{p M E}$ was also used to approximate $\bigwedge$ in Eq.[\ref{eq:loss}]. Further details on the role of $p_{\exists}$ and $p_{\forall}$ can be found in previous works \cite{LTN}.



\textbf{Querying the knowledge base}. 
At inference time, the class with the highest score is selected as the predicted class:

\begin{equation}
\hat{y}=\underset{\tilde{y} \in \mathcal{Y}^U}{argmax} (g(x)^{\mathrm{T}} V a_{\tilde{y}})
\end{equation}

FVLN was evaluated in both ZSL and GZSL settings.  In the ZSL setting, only unseen images are assumed to be present at test time, whereas in the GZSL setting, the model is tested on both seen and unseen classes. This setup induces a bias towards seen classes. To mitigate it, we employed the Calibrated Stacking method, as proposed in \cite{Chen2022MSDNMS,Wang2023GeneralizedZA}, to diminish the classification score of seen classes. The class score is thus calculated as $\hat{y}$:
\begin{equation}
\hat{y}=\underset{\tilde{y} \in \mathcal{Y}^U \cup \mathcal{Y}^S}{argmax }( g(x)^{\mathrm{T}} V a_{\tilde{y}}-\gamma \mathbb{I}\left[\tilde{y} \in \mathcal{Y}^S\right])
\end{equation}
where $\mathbb{I}=1$ if $\tilde{y}$ is from a seen class and zero otherwise, $\gamma$ is a calibration coefficient tuned on a validation set and $\mathcal{Y}^S$ are the labels of seen classes.

\textbf{Construction of the training batch}. Following the approach in \cite{CC-ZSL}, for a positive input image $x_{i}$, we select a set of positive examples $x^{+}$ and $K$ negative examples ${ x_{1}^{-},...,x_{K}^{-} }$. Positive examples are selected from the same category as $x_{i}$, while negative examples are randomly selected from the remaining classes.

\section{Experimental settings}
\label{sec:Experiments}
In this section we examine the datasets used in our experiments, the knowledge base adopted for each dataset and the list of hyperparameters chosen.

\textbf{Dataset}. The experiments were conducted on the AWA2~\cite{awa2}, CUB~\cite{CUB}, and SUN~\cite{SUN} benchmarks. The evaluation metrics for GZSL were based on the standards defined in previous work~\cite{awa2}. 
Following a similar approach outlined in~\cite{Sikka2020ZeroShotLW}, we construct a semantic hierarchy by grouping the classes from Awa2 and CUB datasets into a total of 9 and 49 macroclasses, respectively.


\textbf{Knowledge base}. The knowledge base $\KK$ composition was detailed in Section \ref{sec:flvn}. To build class hierarchy, we map classes from the AWA2 and CUB datasets to their corresponding synsets in WordNet as done by Sikka et al. \cite{Sikka2020ZeroShotLW}. Macroclasses for each dataset were defined by selecting the synset root whose subtree contained classes from the selected dataset, and then defining its immediate children (using WordNet) as classes. Since classes in the SUN dataset~\cite{SUN} lack a semantic hierarchy, axioms related to macroclasses were not included in the knowledge base. For all experiments, in the predicate $\texttt{isofClass}_{masked}$, $k=15$ attributes were randomly dropped. The $\alpha$ parameter in Eq. \ref{eq:gt_sameattributes} was set to 0.01 for AWA2 dataset and 1 on CUB e SUN. To account for the presence of outliers in the knowledge base, we initially set the aggregation function parameters (defined in Eqs. \ref{eq:aggregmean} and \ref{eq:aggregmeanerror})  as $p_{\exists}=2$ and $p_\forall=2$. Both parameters were incremented by 2 every 4 epochs for the AWA2 and CUB datasets; for SUN, the aggregation values were increased at specific epochs (2, 4, 24, and 32) until reaching $p_\exists=6$ and $p_\forall=6$, following the schedule suggested in~\cite{LTN}.

\textbf{Hyperparameter selection}. The embedding function $f_{\theta}$ is based on an ImageNet pre-trained ResNet101 model  that converts the $224 \times 224$ image  into a vector $\mathbf{x} \in \reals^{H \times W \times B}$, where $B=2048$, and $H$ and $W$ represent the height and width of the extracted features; the function $g_\theta$ then converts these features into the attribute space dimensionally consistent with the dataset.

To mitigate overfitting, we initially trained the head with a frozen backbone and subsequently fine-tuned the entire network. We used the Adam optimizer with learning rate of $1e-4$ for the pre-training phase for AWA2 and CUB and 5e-4 for SUN. The learning rate was then reduced to $1e-6\alpha$ for CUB and SUN and $1e-7\alpha$ for AWA2 in the fine-tuning phase, where $\alpha=0.8^{epoch//10}$ and \textit{epoch} is the current epoch out of a total of 300 epochs. Additionally, we incorporated a smoothing factor by multiplying the L2 norm with a scaling factor of 5e-4 for AWA2, 5e-6 for CUB, and 1e-3 for SUN. 

Training batches included positive and negative examples with a ratio of 12 to 12 for AWA2, 12 to 8 for CUB, and 12 to 4 for SUN datasets. At inference time, we set the scaling factor $\gamma$ to adjust the scores obtained for the seen classes to 0.7 for AWA2 and CUB and 0.4 for SUN. In all experiments, we applied a random crop and random flip with 0.5 probability for data augmentation. We implemented the architecture in PyTorch, using the LTNtorch library~\cite{LTNtorch}, and trained it on a single GPU nVidia 2080 Ti. Each experiment was repeated three times to calculate the mean and standard deviation.

\section{Results}
\label{sec:Results}

\begin{table*}[!tb]
\centering
\resizebox{12cm}{!}{%
\begin{tabular}{c|cccc|cccc|cccc}
\hline
&\multicolumn{4}{c}{AWA2}&\multicolumn{4}{c}{CUB} &\multicolumn{4}{c}{SUN} \\
\hline
Model&T1&U&S&H&T1&U&S&H&T1&U&S&H  \\
\hline
PROTO-LTN~\cite{Martone2022PROTOtypicalLT}&67.6   &32.0 &83.7 &46.2&48.8  &20.8 &54.3 &30.0 &60.4  &20.4  &\textbf{36.8}  &26.2 \\
DEM~\cite{dem}&67.1&30.5&\textbf{86.4}&45.1&51.7&19.6&57.9&29.2 &61.9&20.5&34.3&25.6 \\
VSE~\cite{vse}&84.4&45.6&88.7&60.2    &71.9&39.5&68.9&50.2&-&-&-&-\\

TCN~\cite{TCN}&71.2&61.2&65.8&63.4&59.5 &52.6&52.0&52.3&61.5&31.2&37.3&34.0\\

CSNL~\cite{Sikka2020ZeroShotLW} $\dagger$&61.0&-&-&0.0&32.5&-&-&0.7&-&-&-&-\\
\hline

AREN~\cite{AREN}&67.9& 54.7&79.1&64.7& 71.8&63.2&69.0&66.0&60.6&40.3&32.3&35.9 \\
APN~\cite{APN}&68.4 &56.5&78.0&65.5&72.0 &65.3&69.3&67.2&61.6 &41.9&34.0&37.6 \\
AMGML~\cite{Li2021AttributeModulatedGM}&71.7 &56.0&74.6&64.0&70.0 &58.2&55.7&56.9&59.7 &42.0&35.1&38.3 \\
CC-ZSL\cite{CC-ZSL}&68.8 &62.2&\underline{83.1}&71.1&74.3&\underline{66.1}&\underline{73.2}&\underline{69.5}&62.4& 44.4&36.9&40.3 \\

\hline
Cycle-CLSWGAN~\cite{cycle-CLSWGAN}&-&-&-&-&58.4&45.7&61.0&52.3&60.0&\textbf{49.4}&33.6&40.0  \\
LisGan~\cite{LisGAN}&-&-&-&-&58.8&46.5&57.9&51.6&60.0&42.9&37.8&40.2 \\
E-PGN~\cite{E-PGN}&73.4&52.6&83.5&64.6&72.4&52.0&61.1&56.2&-&-&-&- \\
TGMZ~\cite{TGMZ}& \textbf{78.4} &64.1& 77.3&70.1&66.1 &60.3&56.8&58.5&-&-&-&-\\
CEGZSL~\cite{CEGZSL}&70.4&63.1&78.6&70.0&\underline{77.5}&63.9&66.8&65.3&63.3&48.8&\underline{38.6}&\underline{43.1} \\
DFCA-GZSL~\cite{Su2023DualalignedFC}&\underline{74.7}&\underline{66.5}&81.5&\underline{73.3}&\textbf{80.0}&\textbf{70.9}&63.1&66.8&62.6&48.9&\textbf{38.8}&\textbf{43.3}\\



\hline
FLVN $\dagger$&69.8&65.8&82.3&73.1& 71.2 & 62.6 &  83.1 &  71.5 &61.7&48.4&32.7&39.0\\
&\textpm 0.8 &\textpm 0.9&\textpm 0.3&\textpm 0.6& \textpm 0.2 & \textpm  0.5 & \textpm 0.3 & \textpm 0.2 &\textpm0.18&\textpm0.57&\textpm0.2&\textpm 0.1\\
&(71.0)&\textbf({67.1})&(82.8)&(\textbf{74.1})&(71.4)& (63.2)&\textbf({83.4}) & (\textbf{71.7})&(61.9)&\underline({48.9})&(32.9)& (39.1) \\
\hline
\end{tabular}}
\vspace*{5mm}
    \caption{Performance on AWA2, CUB and SUN test set.  For FLVN, we show mean ± standard deviation and maximum (in parentheses) values for TOP1zsl (T1), TOP1gzsl unseen (U), TOP1gzsl seen (S), and Hgzsl (H) over three runs.  Metrics are described in~\cite{awa2}. The first section of the table includes embedding-based models, the second section attention-based models, and the third section generative models. The best performance values are shown in bold. $\dagger$  highlights method which incorporate external knowledge.}
\label{tab:table_test}
\end{table*}
Experimental results presented in Table~\ref{tab:table_test} show how the proposed FLVN architecture reaches competitive performances with respect to other embedding-based methods, particularly CC-ZSL~\cite{CC-ZSL} and APN~\cite{APN}. In general, FVLN performs better in terms of harmonic mean (H) and shows a greater ability to recognize both seen and unseen classes, achieving state-of-the-art performance on two out of three benchmark datasets. In particular, FLVN improves the accuracy of unseen classes and the harmonic mean by 0. 89\% and 1.3\%, respectively, in AWA2, and the accuracy of seen classes and harmonic mean by 12\% and 3\%, respectively, on CUB. These results suggest that the proposed architecture is capable of recognizing seen and unseen samples more evenly, showing less confusion in classifying examples. On SUN, FLVN performs comparably to the state of the art; it is important to note that, in this case, axioms on macro classes are not introduced, therefore FLVN cannot leverage external semantic knowledge. 

Compared to the closest methods in terms of performance, the proposed method requires a relatively simple architecture. Unlike APN, FVLN does not require additional weights for each attribute or supplemental regularization terms to construct the loss function. FVLN only requires a single backbone replica instead of two as in CC-ZSL, which is based on the teacher-student framework~\cite{CC-ZSL}. Compared to generative methods, FLVN does not require to generate additional training samples or make assumptions about unseen classes at train time. 

Although complete ablation studies are left to future work, qualitative observations from our experiments suggest that axioms included in the knowledge base may have quite different impact on the overall performance. With respect to our previous architecture \cite{Martone2022PROTOtypicalLT}, it was crucial to change the  $\texttt{isOfClass}$ grounding for end-to-end training to succeed. The introduction of the $\texttt{isOfMacroclass}$ predicate, on the other hand, introduces smaller improvements. In fact, during training related classes appear to naturally cluster in feature space, as evident from the distribution of the feature space already reported in previous work \cite{Martone2022PROTOtypicalLT}. On the other hand, the $\texttt{isOfClass_{masked}}$ offered important benefits, allowing to account for mistakes in the semantic annotation of classes or traits that exist at the class level but are not visible or easily inferred from a single image (e.g., ``All zebras are agile''). Finally, the \texttt{hasSameAttribute} predicate improved performance particularly on the CUB (fine-grained bird recognition) and SUN (scene recognition) datasets, in which extracting fine-grained image-level attributes is more difficult.

\section{Conclusions and future work}
\label{sec:Conclusion}
Building upon principles from the recent ZSL literature \cite{APN} and incorporating them within a NeSy framework \cite{Martone2022PROTOtypicalLT}, we introduced a novel NeSy architecture, named FLVN, for ZSL and GZSL tasks. FLVN incorporates axioms that combine prior class-specific knowledge (e.g., class hierarchies), with high-level inductive biases to handle exceptions within the dataset (e.g., ``there exists a zebra that is not agile'') and establish relationships between images (e.g., ``if two images belong to the same class, they must be similar''). Such axioms act as semantic priors and compensate for the lack of annotations, thereby providing a solid NeSy foundation for GZSL tasks. FLVN does not require multiple backbones and maintains roughly the same parameter count of standard embedding-based methods. The proposed approach can be also incorporated into other architectures by, e.g., changing the grounding of the predicates. Experimental results prove that FLVN achieves performance on par or exceeding that of current literature on common GZSL benchmarks.  

We believe that our approach can be further extended in several ways. For example, different formulations of the $\texttt{isOfClass}$ predicate, or the introduction of a \texttt{hasAttribute} predicate to predict image-level attributes, could incorporate attention to discriminative regions within the image and increase explainability. Introducing attention mechanisms based on object detection~\cite{manigrasso2021faster} could support the definition useful axioms that identify the most discriminative parts of the image. Introducing axioms that solely consider the similarity between images (with no knowledge of their class membership) could improve results for unseen classes without requiring additional labels in the dataset. The knowledge base could be also extended to consider other types of relationships, beyond class hierarchies, and from emerging sources, such as language models.  Finally, the proposed method could be explored in a trasductive ZSL setting.

\section*{Acknowledgements}
This study was carried out within the FAIR - Future Artificial Intelligence Research and received funding from the European Union Next-GenerationEU (PIANO NAZIONALE DI RIPRESA E RESILIENZA (PNRR) – MISSIONE 4 COMPONENTE 2, INVESTIMENTO 1.3 – D.D. 1555 11/10/2022, PE00000013). This manuscript reflects only the authors’ views and opinions, neither the European Union nor the European Commission can be considered responsible for them.

\bibliographystyle{splncs05}
\bibliography{main.bib}

\end{document}